\documentclass[conference]{IEEEtran}
\IEEEoverridecommandlockouts
\usepackage{cite}
\usepackage{amsmath,amssymb,amsfonts}
\usepackage{algorithmic}
\usepackage{graphicx}
\usepackage{textcomp}
\usepackage{xcolor}
\usepackage{graphicx}
\usepackage{subcaption}
\usepackage{hyperref}
\usepackage{authblk}

\graphicspath{{./Figures}}

\def\BibTeX{{\rm B\kern-.05em{\sc i\kern-.025em b}\kern-.08em
    T\kern-.1667em\lower.7ex\hbox{E}\kern-.125emX}}
\begin{document}

\title{Fighter Jet Navigation and Combat using Deep Reinforcement Learning with Explainable AI
}

\author[1]{Swati Kar}
\author[2]{Soumyabrata Dey}
\author[2]{Mahesh K Banavar}
\author[1]{Shahnewaz Karim Sakib}
\affil[1]{Dept. of CS, University of Tennessee, Chattanooga}
\affil[2]{Dept. of ECE, Clarkson University, Potsdam, NY}

\maketitle

\begin{abstract}
    This paper presents the development of an Artificial Intelligence (AI) based fighter jet agent within a customized Pygame simulation environment, designed to solve multi-objective tasks via deep reinforcement learning (DRL). The jet's primary objectives include efficiently navigating the environment, reaching a target, and selectively engaging or evading an enemy. A reward function balances these goals while optimized hyperparameters enhance learning efficiency. Results show more than 80\% task completion rate, demonstrating effective decision-making. To enhance transparency, the jet's action choices are analyzed by comparing the rewards of the actual chosen action (factual action) with those of alternate actions (counterfactual actions), providing insights into the decision-making rationale. This study illustrates DRL's potential for multi-objective problem-solving with explainable AI. Project page is available at: \href{https://github.com/swatikar95/Autonomous-Fighter-Jet-Navigation-and-Combat}{Project GitHub Link}.

\end{abstract}

\begin{IEEEkeywords}
    Artificial Intelligence, Deep Reinforcement Learning, Multi-Objective, Pygame Simulation, Explainability, Reward Function Analysis
\end{IEEEkeywords}

\section{Introduction}
In recent years, rapid advancements in technology have positioned AI as a transformative force across various fields. AI's ability to emulate human intelligence has led to groundbreaking developments, reshaping industries and redefining how to approach complex tasks. Traditionally, many critical tasks relied heavily on skilled human intervention, especially in high-risk situations. However, AI now has the potential to autonomously undertake these tasks, reducing risk to human life and improving operational efficiency. Starting with notable achievements, such as surpassing human capabilities in chess in 1997, AI has expanded to tackling the highly intricate board game Go \cite{feng-hsiunghsuIBMsDeepBlue1999,silverMasteringGameGo2016}. This shift highlights a new era where AI not only matches but often surpasses human abilities in executing high-stakes, strategic tasks. RL, a subset of AI, enables an agent to learn effective actions within its environment through trial-and-error interactions, eliminating the need for human expert data to identify robust Courses of Action (CoAs) \cite{selmonajExplainabilityMultiAgent}. 

In the field of jet navigation and combat, several previous works have been conducted. The simulation model, as explored in~ \cite{piaoVisualRangeAirCombat2020}, primarily focuses on air combat scenarios rather than on reinforcement learning strategies or explainability. It lacks the detailed DRL approach and reward function design, and consequently, the emphasis is more on simulating combat than on optimizing learning strategies. Reinforcement learning is covered in~\cite{huangInfraredAirCombat2021}, though it does not delve into explainability through factual versus counterfactual analysis. Instead, the focus lies on the technical aspects, such as network architecture and training processes, rather than on the decision-making transparency of the agent. Additionally,~\cite{huangInfraredAirCombat2021} lacks a detailed account of how agents improve their efficiency in task completion over time.The simulation-based pilot training systems discussed in~\cite{kallstromDesignSimulationbasedPilot2022} focus more on training scenarios and the roles of agents rather than providing an in-depth DRL approach. The reward systems discussed in~\cite{kallstromDesignSimulationbasedPilot2022} align more with training goals than the complex balance of efficiency and resource management featured in our proposed approach. Simpler reward functions, as noted in~\cite{baeDeepReinforcementLearningBased2023}, typically concentrate on immediate task outcomes such as shooting down targets or avoiding crashes. While these reward functions effectively guide the agent’s learning, they lack the complexity and balance necessary for encouraging nuanced decision-making processes, which require balancing long-term efficiency with short-term actions. 

In this research, all of these shortcomings are addressed. Therefore, the main contributions of this paper are:

\begin{enumerate}
    \item A reward schema that balances efficiency, resource management, and intelligent decision-making to address a multi-objective problem.
    \item Enhanced explainability through factual and counterfactual analysis, providing insights into the agent’s decisions and improving transparency.
\end{enumerate}

\begin{figure*}[!t]
    \centering
    \includegraphics[width=0.85\textwidth]{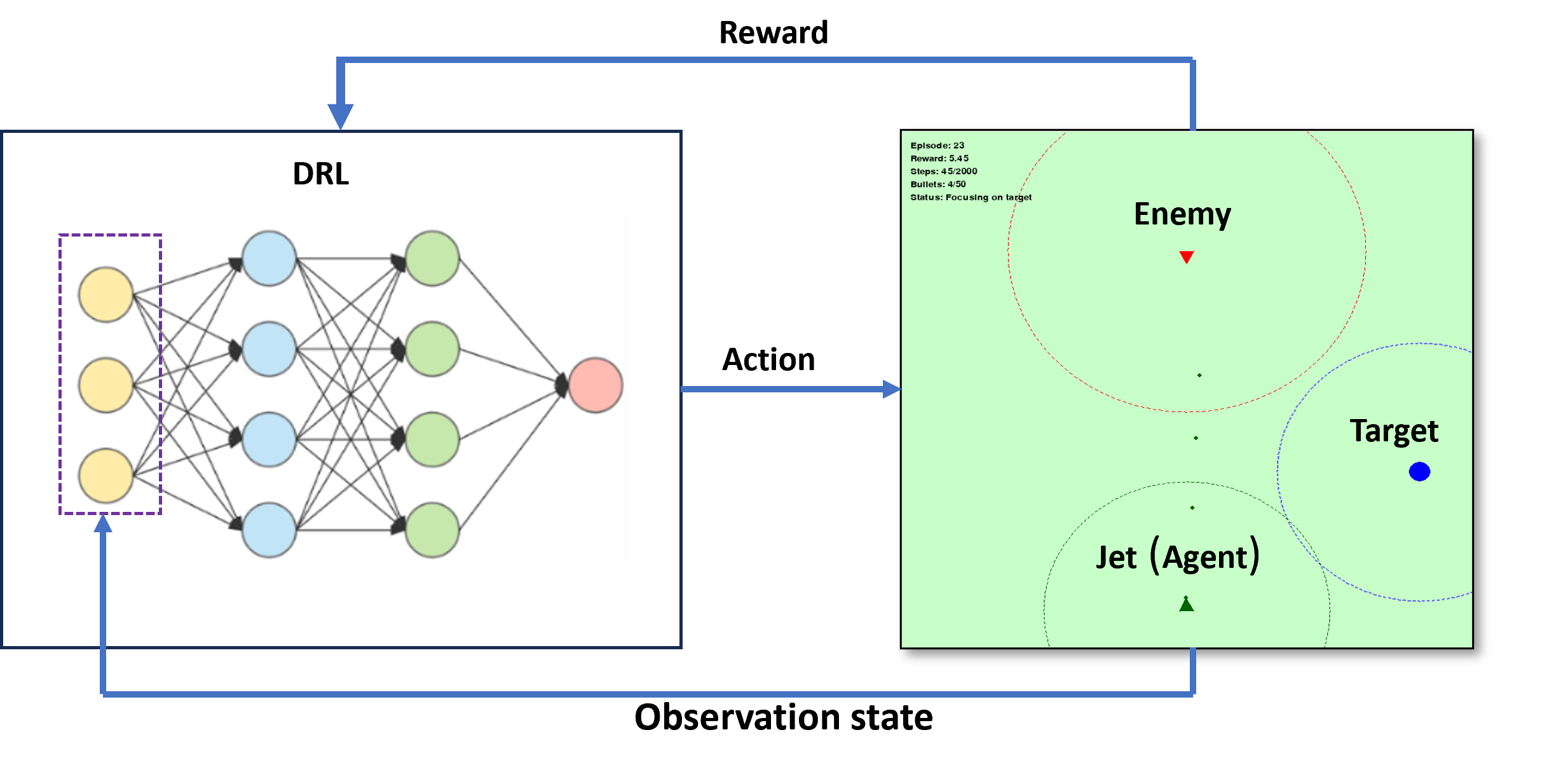}
    \caption{Workflow between DRL and Fighter jet environment}
    \label{fig:pygame_env}
\end{figure*}

Our study is organized into the following key components: first, we develop a custom simulation environment. Next, we train a fighter jet agent to make strategic engagement decisions using the double deep q-learning (DDQN) algorithm. We then focus on optimizing mission resources and explaining the agent's decision-making process through factual and counterfactual scenarios. By addressing challenges such as prioritization, adaptive behavior, and risk assessment, this research aims to advance the development of intelligent, autonomous systems for complex, multi-objective scenarios, ultimately enhancing AI's role in high-stakes environments.


\section{Fighter Jet Problem Formulation}

\subsection{Environment Design}

\begin{table}[!h]
\centering
\caption{Target and Object Representations in the Simulation}
\renewcommand{\arraystretch}{1.2}
\begin{tabular}{p{3cm} p{4cm}}

\hline
\textbf{Object} & \textbf{Representation} \\
\hline
Target & Blue circle \\
Target Range & Blue dotted circle \\
Agent & Green triangle \\
Agent Range & Green dotted circle \\
Enemy & Red triangle \\
Enemy Range & Red dotted circle \\
\hline
\end{tabular}
\label{tab:target_representation}
\end{table}

The high level simulation environment with reinforcement learning agent is depicted in Figure~\ref{fig:pygame_env}. This is a continuation of previous research~\cite{karComparisonExplainabilityApproaches2024} The simulation environment is designed using Python's Pygame package. In the visualization, the blue circle represents the target, the red triangle is the enemy jet and the green triangle is the agent jet as described in Table~\ref{tab:target_representation}. The design of the environment incorporates the following constraints to enhance the realism and challenge for the reinforcement learning agent:

\begin{enumerate}
    \item \textbf{Targeting Zone Constraint}: 
    The agent can only successfully hit the target by firing within a designated blue targeting zone surrounding the target. 
    This constraint encourages the agent to strategically position itself before attacking.

    \item \textbf{Limited Observation Ranges}: 
    Both the agent and the enemy have restricted observation ranges. The enemy does not react to the agent if it is outside the red radius, and the agent cannot observe the enemy if it is outside the green sphere. 
    These constraints simulate realistic sensory limitations and require the agent to operate within a limited field of view.

    \item \textbf{Environment Boundary Constraint}: 
    Any moves that would cause the agent to exit the visible environment area are nullified, leaving the agent in place. 
    This constraint ensures the agent remains within the simulation boundaries, preventing unintended behavior outside the defined environment.
\end{enumerate}

 The jet's position \(\mathbf{p_{jet}} = [x_{jet}, y_{jet}]\) is updated based on its velocity vector \(\mathbf{v_{jet}}\), which is a function of the speed \(v\) and orientation \(\theta_{jet}\):
 \begin{equation}
    \mathbf{v_{jet}} = v \begin{bmatrix} \cos(\theta_{jet}) \\ \sin(\theta_{jet}) \end{bmatrix},
 \end{equation}
 \begin{equation}
    \quad \mathbf{p_{jet}}(t+1) = \mathbf{p_{jet}}(t) + \mathbf{v_{jet}} \Delta t.
 \end{equation}

\noindent
The orientation \(\theta_{jet}\) changes according to a turn rate \(\Delta \theta\), depending on the chosen action (left or right turn).
The jet's speed \(v\) adjusts based on acceleration or deceleration inputs:
\begin{equation}
    v_{t+1} = \max(\min(v(t) + \alpha \Delta t, v_{\text{max}}), v_{\text{min}}) ,
\end{equation}
    
\noindent
ensuring the speed remains within specified limits.
The Euclidean distance \(d_{target}\) between the jet and the target is:
\begin{equation}
    d_{target} = \sqrt{(x_{jet} - x_{target})^2 + (y_{jet} - y_{target})^2}.
\end{equation}

\noindent
Similarly, the angle \(\theta_{target}\) between the jet and the target can be calculated as:
\begin{equation}
    \theta_{target} = \tan^{-1} \left( \frac{y_{target} - y_{jet}}{x_{target} - x_{jet}} \right).
\end{equation}

\noindent
Upon firing, the bullet's velocity vector \(\mathbf{v_{bullet}}\) is defined in the direction of \(\theta_{jet}\), and its position updates as follows:
\begin{equation}
    \mathbf{p_{bullet}}(t+1) = \mathbf{p_{bullet}}(t) + \mathbf{v_{bullet}}.
\end{equation}

\noindent
Collision detection is based on the proximity of the bullet to the target or enemy, with a threshold to confirm a hit.

\subsection{State Space}

The state variables are chosen to provide the fighter jet a comprehensive understanding of both its own status and its interactions with key elements in the environment, enabling informed decision-making for effective navigation and engagement. The jet's state space consists of positional, directional, and interaction variables: global coordinates $(x_{jet}, y_{jet})$, heading angle $\theta_{jet}$, and velocities $(v_x, v_y)$. Relative measurements include the alignment angle $\alpha_{target}$ and distance $d_{target}$ to the target, as well as the angle $\alpha_{enemy}$ and distance $d_{enemy}$ to the enemy. Visibility indicators include $V_e$ (enemy visibility), $V_b$ (bullet visibility), and $T_z$ (target zone presence). Additionally, $d_b$ represents the distance between the agent and the nearest bullet. The complete state vector is represented as:

\begin{equation}
    \begin{aligned}
        S_i &= \bigl[
        x_{\mathrm{jet}},\, y_{\mathrm{jet}},\, \theta_{\mathrm{jet}},\, v_x,\, v_y,\,
        \alpha_{\mathrm{target}},\, d_{\mathrm{target}},\\
        &\quad \alpha_{\mathrm{enemy}},\, d_{\mathrm{enemy}},\, V_e,\, V_b,\, T_z,\, d_b
        \bigr]^\top
    \end{aligned}
\end{equation}


\begin{figure} 
    \centering
    \includegraphics[width=0.5\textwidth]{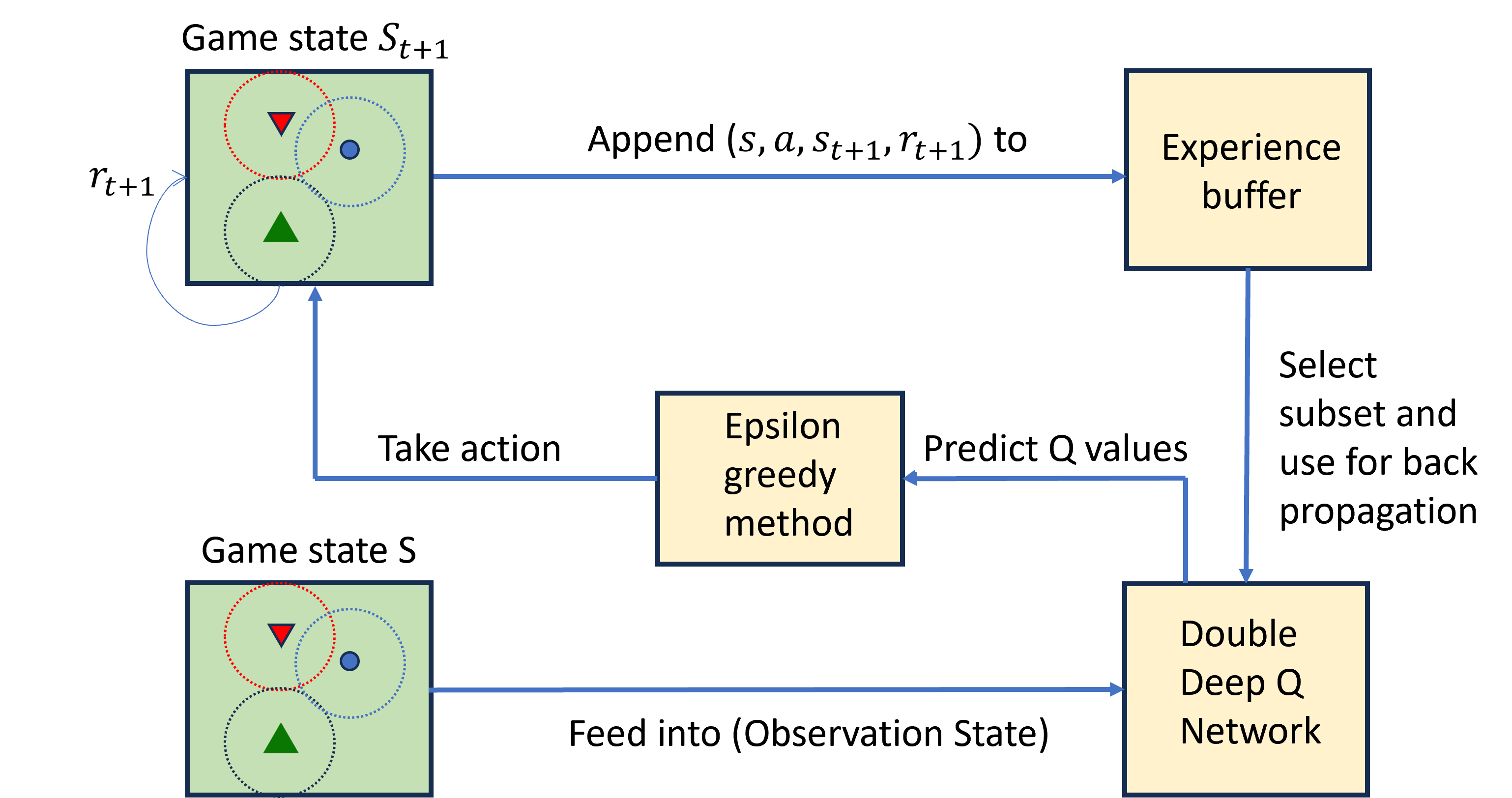}
    \caption{Fighter Jet Double DQN Workflow}
    \label{fig:dqn_workflow}
\end{figure}

\begin{figure} 
    \centering
    \includegraphics[width=0.45\textwidth]{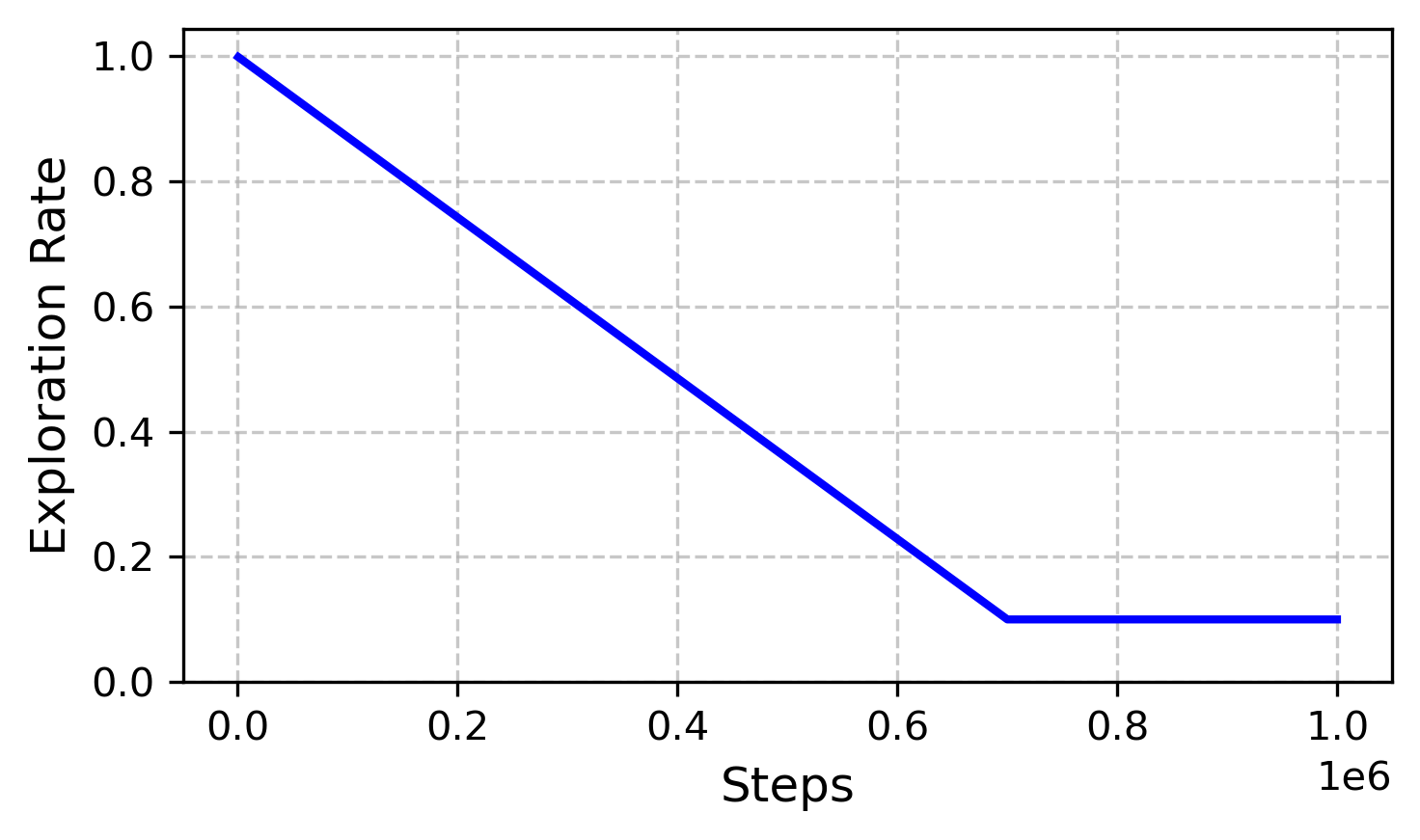}
    \caption{Epsilon Decay over Time}
    \label{fig:epsilon}
\end{figure}

\subsection{Action Space}

The agent's action space consists of six discrete actions: $a_0$ maintains the current orientation and speed, $a_1$ and $a_2$ adjust the heading angle by turning left or right by $0.05$ radians, respectively. The actions $a_3$ and $a_4$ control the speed by accelerating or decelerating by $0.25$ units, and $a_5$ enables the agent to shoot, generating a new bullet. The continuous actions are discretized to simplify the agent's decision-making process and improve explainability, allowing clear interpretation of the agent’s choices and their impact on the environment. The complete action space vector is defined as:

\[
A = [a_0, a_1, a_2, a_3, a_4, a_5]^\top
\]

\subsection{Reward Function}

The reward function balances task performance, efficiency, and resource management by incorporating rewards for desirable actions and penalties for inefficiencies. It was designed through a trial-and-error strategy to fine-tune the agent’s behavior, encouraging timely task completion and effective interaction with the environment. Positive rewards are provided for proximity to the target, target zone alignment, and successful hits, while penalties discourage resource wastage, missed opportunities, and mission failures. The reward function is defined as:
\begin{align}
    R = & -0.1 - 15 \cdot \max(0, d_{cur} - d_{prev}) - (1 - I_{tz}) - 0.5(1 - I_{ez}) \nonumber \\
    & - 0.5 \cdot \max (0, b - 50) - 500 \cdot I_{eh} - 1000 \cdot I_{ms}\nonumber \\
    & + 10(d_{prev} - d_{cur}) + 2I_{tz} + I_{ez} \nonumber \\
    & + 200 \cdot I_{ht} + 100 \cdot I_{he}\nonumber,
\end{align}

\noindent
where,
\begin{itemize}
    \item $d_{\text{prev}}, d_{\text{curr}}$: Previous and current distances to the target.
    \item $I_{tz}$: Indicator for target zone presence.
    \item $I_{ez}$: Indicator for spotting the enemy.
    \item $b$: Current bullet count.
    \item $I_{\text{ht}}$: Indicator for hitting the target.
    \item $I_{\text{he}}$: Indicator for hitting the enemy.
    \item $I_{\text{ms}}$: Indicator for mission failure.
\end{itemize}

\subsection{Training and Testing Analysis of the DDQN Algorithm}

This study employs the double deep Q-network (DDQN) ~\cite{vanhasseltDeepReinforcementLearning2016} algorithm for decision-making in a fighter jet simulation environment shown in Figure~\ref{fig:dqn_workflow}. In Q-learning, the Q-value represents the expected cumulative reward for taking an action in a given state and following the optimal policy thereafter. Traditional deep Q-networks (DQN) often overestimate Q-values due to the use of the same network for action selection and Q-value estimation, which can lead to suboptimal policies. DDQN mitigates this issue by decoupling these tasks, using the target network for Q-value estimation, resulting in more stable and accurate learning. 

The neural network architecture for the Q-network consists of three fully connected layers with 256 neurons in each layer. The ReLU activation function introduces non-linearity to enhance learning capability, while the Adam optimizer is utilized for efficient gradient descent and weight updates. Hyperparameters used for training are shown in Table ~\ref{table:ddqn_hyperparameters} which were achieved through trial and error.

The epsilon decay policy balances exploration and exploitation in the double DQN algorithm. Exploration means trying random actions to gather information, whereas exploitation means using the best-known actions to maximize reward. Starting with an epsilon value of $1.0$, the agent explores the environment by selecting random actions. Epsilon decreases linearly over time, reaching a minimum of $0.1$, encouraging the agent to shift from exploration to exploitation, leveraging learned knowledge for decision-making. At $0.1$, the agent primarily selects optimal actions while occasionally exploring to refine its policy. This gradual transition enhances stability and performance, ensuring a smooth shift from exploratory learning to policy-driven actions. The linear decay pattern is illustrated in Figure~\ref{fig:epsilon} for training steps.
\begin{table}[!t]
    \centering
    \caption{DDQN Hyperparameters}
    \renewcommand{\arraystretch}{1.2}
    \begin{tabular}{p{3cm} p{3.5cm}}

    \hline
    \textbf{Hyperparameter}                  & \textbf{Value}  \\ \hline
    Learning Rate                            & 0.00005         \\
    Discount Factor (\(\gamma\))             & 0.99            \\
    Replay Buffer Size                       & 500,000         \\
    Batch Size                               & 256             \\
    Target Update Interval                   & 5000 steps      \\
    Exploration Initial Epsilon              & 1.0             \\
    Exploration Final Epsilon                & 0.1             \\
    Exploration Fraction                     & 0.7             \\
    Max Gradient Norm                        & 10              \\
    Network Architecture                     & [256, 256, 256] neurons \\
    \hline
    \end{tabular}
    \label{table:ddqn_hyperparameters}
\end{table}

\begin{figure*}[!t]
	\centering
	\begin{minipage}{0.48\textwidth}
        \centering
        \includegraphics[width=1\textwidth]{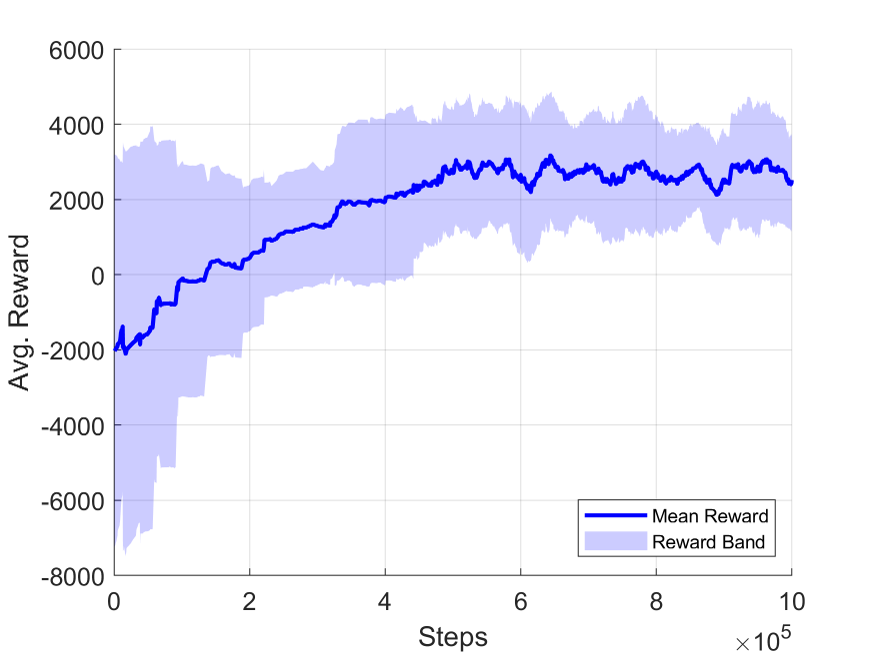}
        \caption{Average Reward vs Steps}
        \label{fig:avg_reward}
	\end{minipage}\hfill
	\begin{minipage}{0.48\textwidth}
        \centering
        \includegraphics[width=1\textwidth]{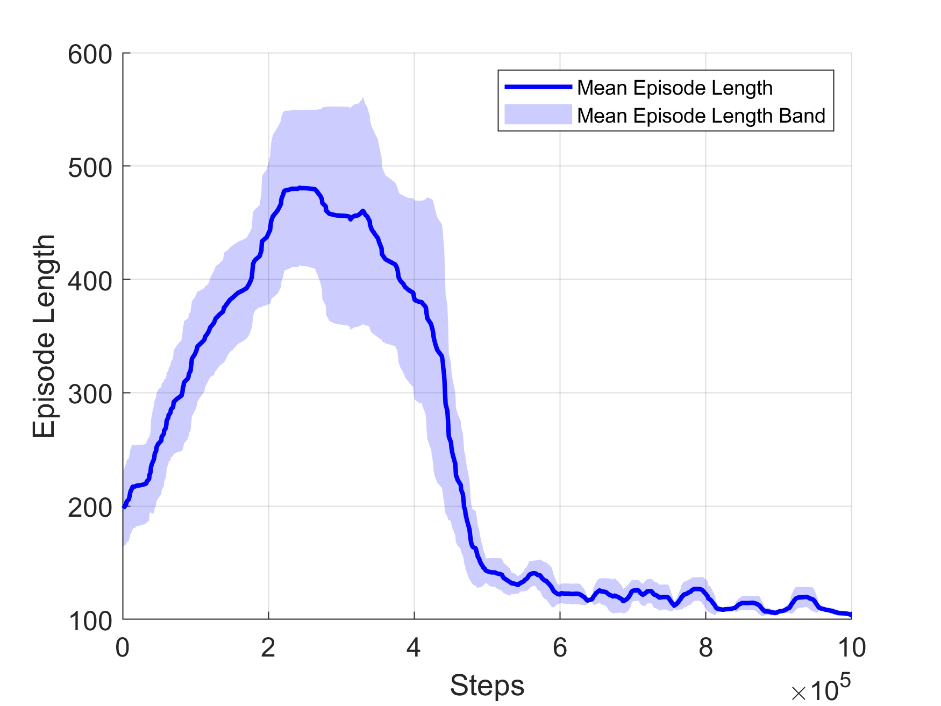}
        \caption{Mean Episode Length vs Steps}
        \label{fig:episode_length}
	\end{minipage}
\end{figure*}

\section{\textbf{Result}}
The agent's training performance is evaluated based on key metrics, including episode length, average reward, success rate, and trajectory analysis. These metrics provide insights into the agent's learning progress, decision-making efficiency, and task completion capabilities, validating the effectiveness of the DDQN algorithm in the fighter jet simulation environment. The details about step, episode and episode termination conditions are shown in Table~\ref{tab:episode_details}. The explainability analysis further enhances transparency by examining the agent's action choices through factual and counterfactual reward comparisons, offering insights into the decision-making process and the rationale behind the agent's actions.

\begin{table}[!h]
    \centering
    \caption{Step and Episode Details with Termination Conditions.}
    \label{tab:episode_details}
    \renewcommand{\arraystretch}{1.2}
    \begin{tabular}{p{2cm} p{5cm}}
    \hline
    \textbf{Criterion}                 & \textbf{Details}                                                \\ \hline
    Step                     & A single iteration where the agent selects and executes an action \\ 
    Episode                     & A complete sequence of interactions between the agent and the environment, 
    \\ & starting from an initial state and ending 
    \\ & at a terminal state                                          \\ 
    Episode Termination Condition      & 1. If the enemy shoots the agent                                 \\ 
                                       & 2. If the agent shoots the target                            \\ 
                                       & 3. If the task is not completed within 2000 steps             \\ \hline
    \end{tabular}
    
\end{table}

\subsection{Fighter Jet Agent Training Performance}

Figure~\ref{fig:avg_reward} illustrates the agent's average reward over 1 million training steps. The average reward increases steadily, reflecting the agent's improvement in performance as training advances. The reward increases steadily, reflecting the agent's ability to learn and apply effective strategies to maximize cumulative rewards. The shaded region shows the reward variance, which reduces over time, indicating that the agent’s actions are becoming more consistent and focused on optimal strategies. This positive trend in average reward validates the effectiveness of the training process, as the agent learns to make decisions that align with the objectives of the task.

Figure~\ref{fig:episode_length} shows the mean episode length over training steps. Initially, the episode length is small as enemy destroyes agent. After the agent learns dealing with enemy, episode length increases gradually. As training progresses, the episode length finally decreases and stabilizes, suggesting that the agent becomes more efficient in achieving its objectives and completing tasks within fewer steps.

\subsection{Success vs Failure Rates}

\begin{table}[!h]
    \centering
    \caption{Success vs Failure Rates (1000 Episodes)}
    \renewcommand{\arraystretch}{1.2}
    \begin{tabular}{p{2cm} p{2.5cm} p{2.5cm}}
    \hline
    \textbf{Outcome} & \textbf{Number of Episodes} & \textbf{Rate (\%)} \\ \hline
    Successes        & 825                         & 82.5               \\
    Failures         & 175                         & 17.5               \\
    \hline
    \end{tabular}
    \label{table:success_failure_rate}
\end{table}

After training, evaluation episodes are conducted to assess the agent's performance in the fighter jet simulation environment. Table \ref{table:success_failure_rate} displays the agent’s success and failure rates across episodes. Out of 1000 testing episodes, the agent successfully completes 825 episodes, achieving the objective, while it fails in 175 episodes. This success rate demonstrates the effectiveness of the DDQN algorithm in guiding the agent toward its goals.

\begin{figure}[!t]
    \centering
    \includegraphics[width=0.5\textwidth]{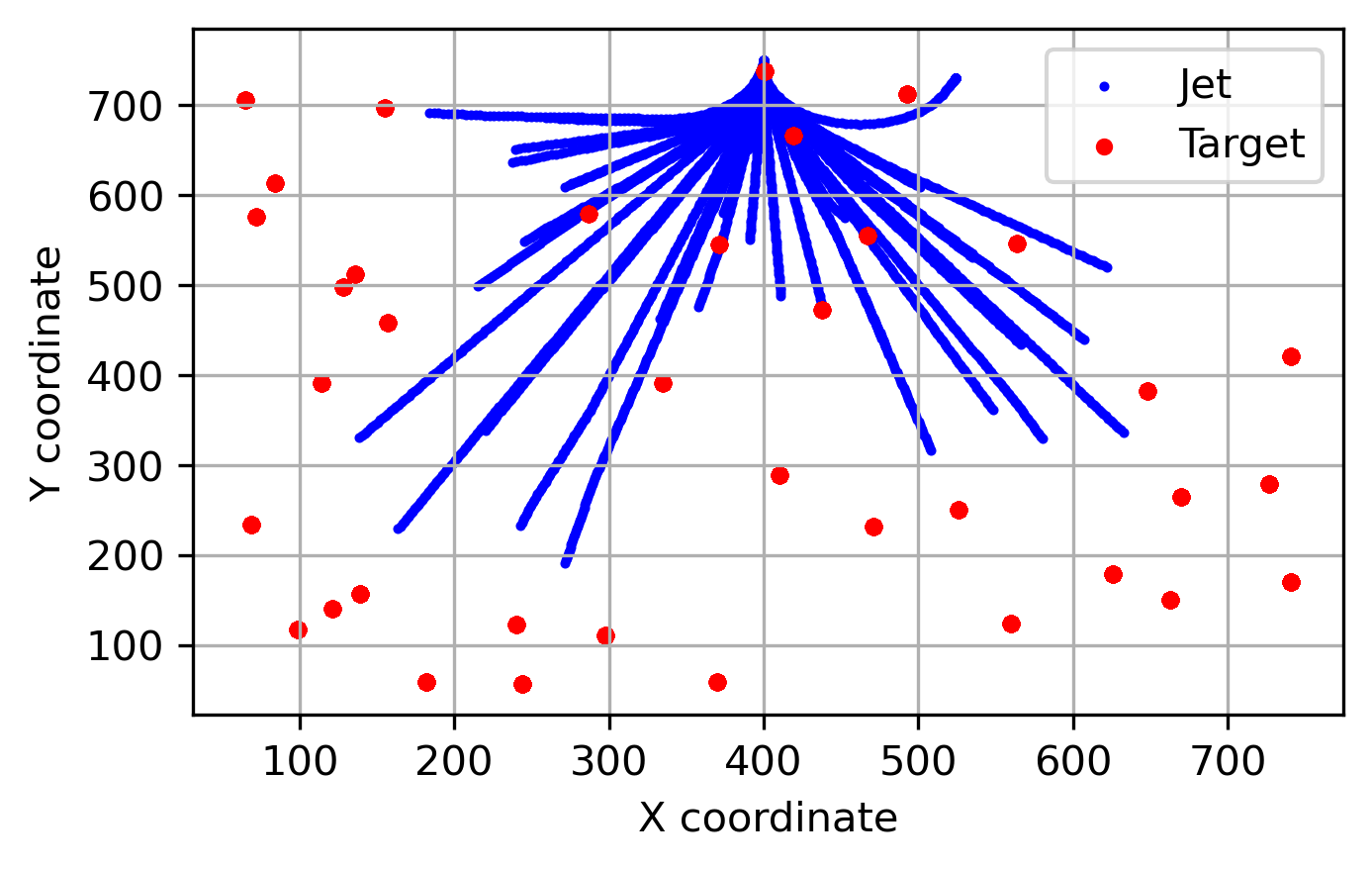}
    \caption{Trajectory of jet collected for 50 episodes}
    \label{fig:trajectory}
\end{figure}

\subsection{Jet and Target Positions}  

The trajectory plot in Figure~\ref{fig:trajectory} illustrates the jet's paths (blue) and target positions (red) over 50 evaluation episodes. The plot highlights the jet's ability to navigate toward various random target position, adapting its trajectory based on the target’s location.  To generate these results, target positions are randomly assigned at the start of each episode, creating diverse scenarios for the agent. The agent collects samples by navigating through the environment and engaging with the target. For simplicity, enemy trajectories are omitted.

The trajectories confirm the agent's efficiency in reaching the target, maintaining proximity, and engaging effectively, validating the DDQN algorithm's performance in multi-objective decision-making tasks. The distance between the jet's final position and the target indicates that once the target enters the jet's targeting zone, the agent engages and concludes the episode.

\section{\textbf{Explainability of Fighter Jet Agent Decisions}}

Explainability in machine learning refers to the ability to convey how a model makes its decisions to a human, even one without technical expertise. This is crucial for building trust and identifying potential biases. In reinforcement learning, explainability is especially vital since the agent makes real-time decisions, and understanding its rationale is key. This section explores techniques for interpreting the decisions of the fighter jet agent in the simulation environment.

\begin{figure*}[!t]
    \centering
    \includegraphics[width=0.95\textwidth]{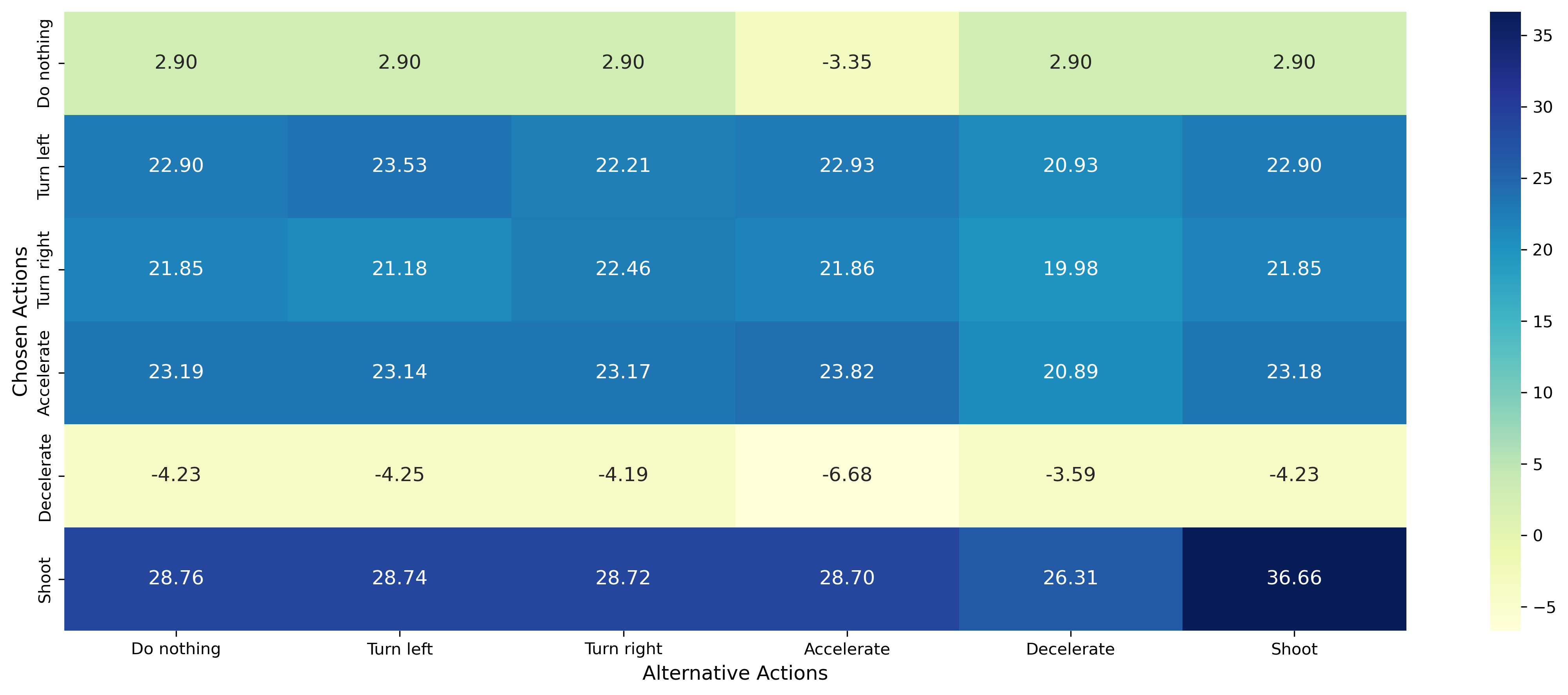}
    \caption{Reward Heatmap: Comparison of Factual and Counterfactual Actions. Each cell shows the average reward for 1000 evaluation episodes, with color intensity representing reward magnitude. Darker shades indicate higher rewards. Diagonal elements correspond to factual actions.}
    \label{fig:reward_heatmap}
\end{figure*}

\subsection{Reward Heatmap: Factual vs Counterfactual Actions}
The reward heatmap in Figure~\ref{fig:reward_heatmap} analyzes the agent's decision-making by comparing factual and counterfactual rewards. Diagonal values represent chosen actions (factual), and others display alternative actions (counterfactual). Each cell shows the average reward for 1000 evaluation episodes if the counterfactual action were taken. The color intensity indicates reward magnitude, with darker shades for higher rewards and lighter shades for lower rewards. Diagonal elements correspond to the agent's factual actions.

From the heatmap, it can be observed that certain actions consistently yield higher rewards, which guides the agent's preferred choices. For instance, when the agent chooses the “Shoot” action, the factual reward is higher than the counterfactual rewards for other actions, indicating that shooting was indeed the optimal choice in those scenarios. Conversely, actions like ``Decelerate" yield lower rewards, suggesting they are less favorable in the given environment conditions.

\subsection{Factual vs Average Counterfactual Rewards}

\begin{figure}[!h]
    \centering
    \includegraphics[width=0.45\textwidth]{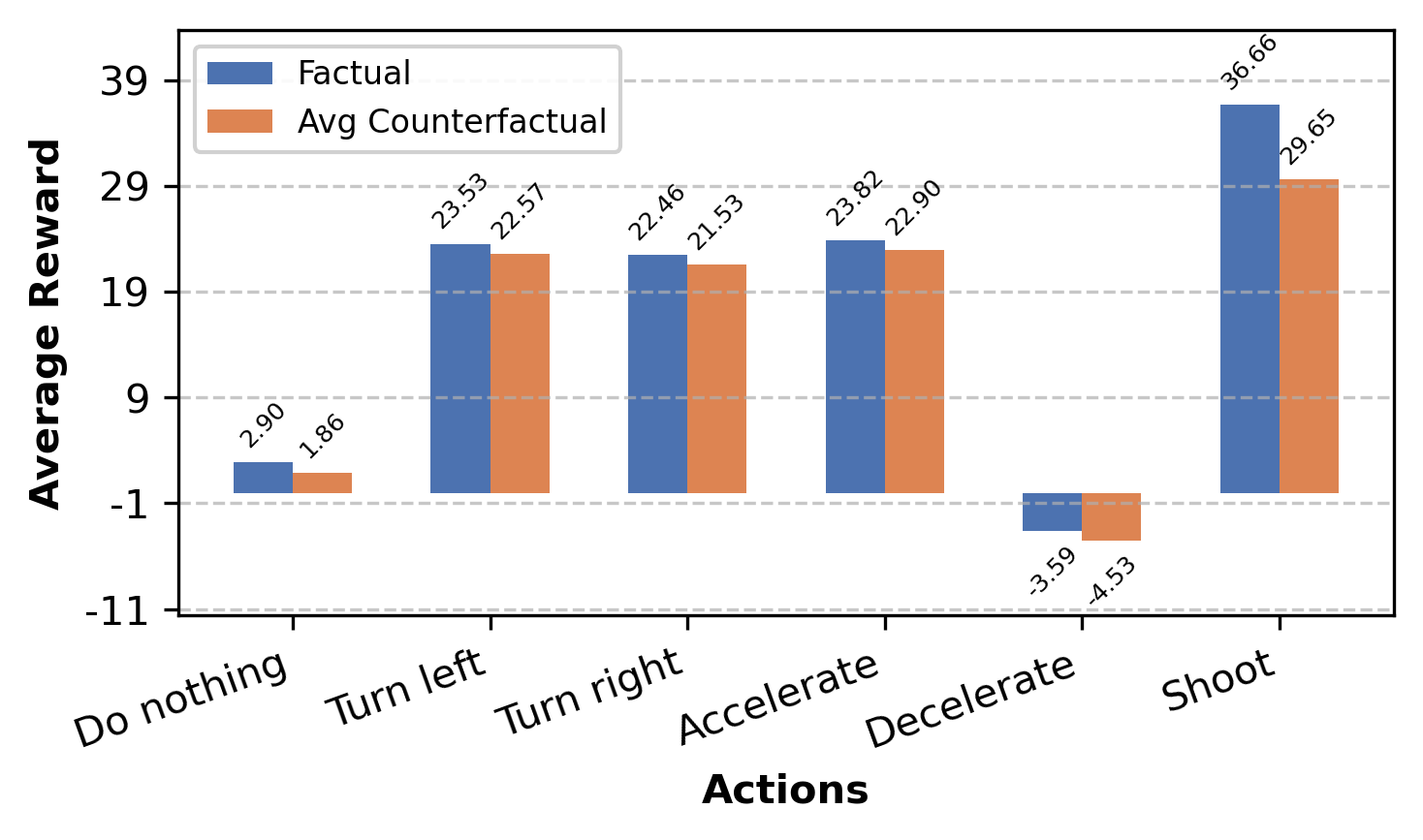}
    \caption{Factual vs Average Counterfactual Rewards}
    \label{fig:avg_rewards}
\end{figure}

Figure~\ref{fig:avg_rewards} compares the average factual rewards for each action with the average counterfactual rewards if alternative actions were taken. The factual rewards, represented by the blue bars, demonstrate the actual rewards the agent received for each chosen action. In contrast, the orange bars represent the average reward the agent would have received if it had selected a different action.

The comparison shows that actions like ``Shoot" and ``Turn Left" consistently provide higher factual rewards than their counterfactual counterparts, reinforcing that these choices were optimal given the state conditions.

\subsection{Distribution of Chosen Actions}

\begin{figure}[!h]
    \centering
    \includegraphics[width=0.45\textwidth]{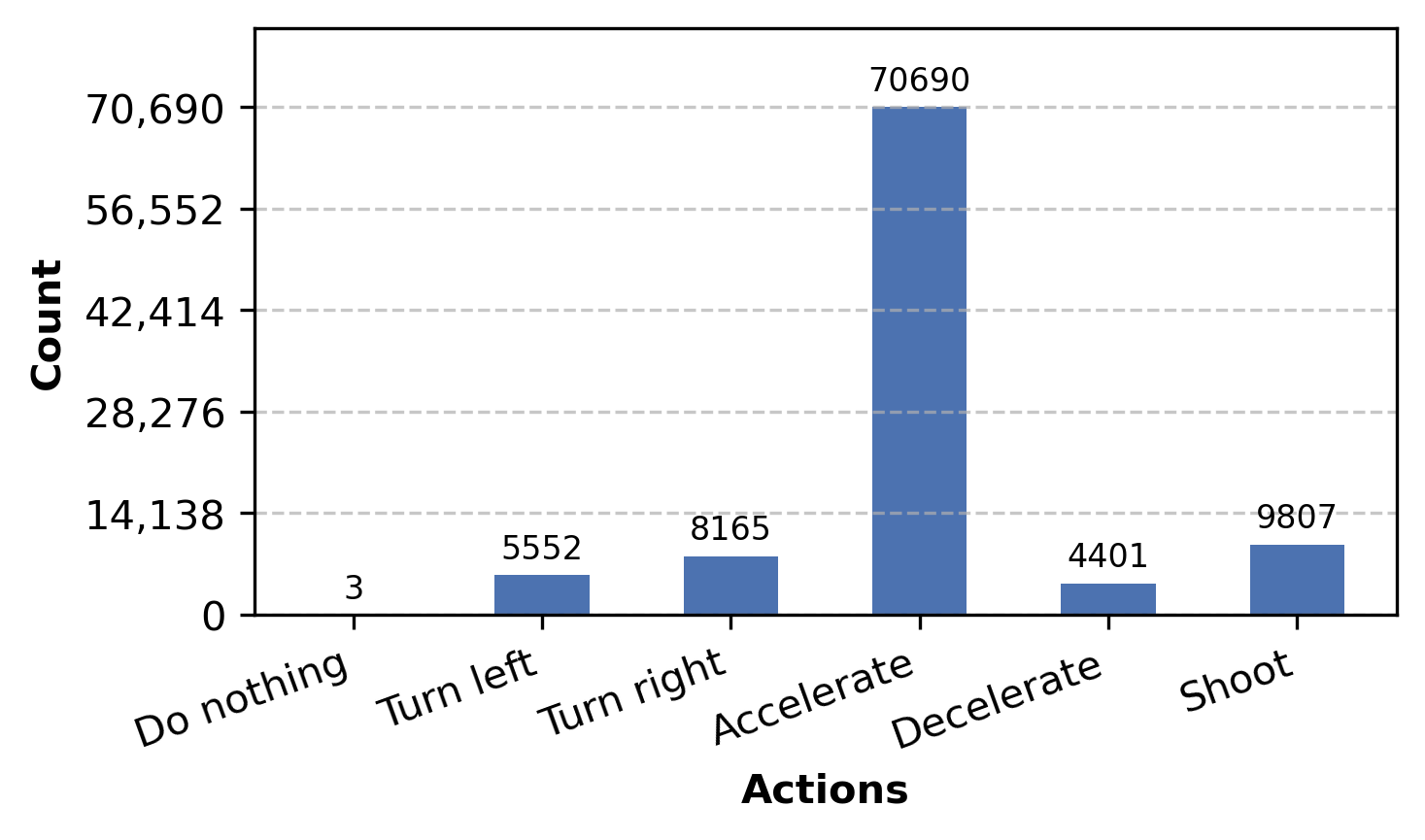}
    \caption{Distribution of Chosen Actions}
    \label{fig:chosen_actions}
\end{figure}

The action distribution plot in Figure~\ref{fig:chosen_actions} shows the frequency of each action selected by the agent during training. The actions “Accelerate” and “Shoot” were chosen most frequently, consistent with their higher factual rewards observed in the previous analysis. In contrast, the “Decelerate” action appeared much less often, corroborating its lower reward outcomes as reflected in the reward heatmap and counterfactual reward comparisons. The lower favorability of “Decelerate” can be attributed to its high risk: when the agent decelerates, it becomes vulnerable to enemy attacks, resulting in frequent defeats and reduced reward accumulation.

This distribution provides an intuitive explanation of the agent's strategy. The agent learns to favor actions with higher reward potentials and reduces the selection of actions with lower rewards. By analyzing the chosen actions alongside factual and counterfactual rewards, we gain insight into the agent's learned policy and the rationale behind its decision-making.

\section{Conclusion and Future Work}

This study demonstrates the effectiveness of a Double Deep Q-Learning (DDQN) algorithm for multi-objective decision-making in a simulated fighter jet environment. The agent achieves a high success rate, effectively balancing navigation, engagement, and resource management. Explainability techniques, such as factual and counterfactual reward analysis, provide valuable insights into the agent’s decision-making process, enhancing transparency and trust.

Future work will focus on extending the simulation to incorporate more complex combat scenarios and multi-agent interactions. Additional efforts will explore integrating advanced explainability methods to further improve the interpretability of agent decisions and adapt the framework for real-world autonomous systems.

\bibliographystyle{IEEEtran}
\bibliography{reference}

\vspace{12pt}

\end{document}